%% file: acl_latex.tex
\documentclass[11pt]{article}

\usepackage[]{acl}

\usepackage{times}
\usepackage{latexsym}

\usepackage[T1]{fontenc}
\usepackage[utf8]{inputenc}

\usepackage{microtype}

\usepackage{inconsolata}

\usepackage{graphicx}
\usepackage[utf8]{inputenc}
\usepackage[linesnumbered,ruled,vlined]{algorithm2e}
\usepackage{amsmath}

\input{commands.tex}

\title{Agentic CLEAR: Automating Multi-Level Evaluation of LLM Agents}

\author{
  \textbf{Asaf Yehudai\textsuperscript{I}}\thanks{Equal contribution.},
  \textbf{Lilach Eden\textsuperscript{I}}\footnotemark[1],
  \textbf{Michal Shmueli-Scheuer\textsuperscript{I}}
\\
\\
  \textsuperscript{I}IBM Research
\\
Asaf.Yehudai@ibm.com,
\{lilache, shmueli\}@il.ibm.com \\
}

\begin{document}
\maketitle
\begin{abstract}
Agentic systems are becoming more capable: agents define strategies, take actions, and interact with different environments. This autonomy poses serious challenges for overseeing and assessing agent behavior. Most current tools are limited, focusing on observability with basic evaluation capabilities or imposing static, hand-crafted error taxonomies that cannot adapt to new domains. To address this gap, we present \method{}, an automatic, dynamic, and easy-to-use evaluation framework. It produces textual insights into the agent behavior on three levels of granularity: system, trace, and node. 
\method{} operates above the observability layer, enabling seamless integration and featuring an intuitive UI that makes agent evaluation highly accessible.
In our experiments on four benchmarks, seven agentic settings, and tens of thousands of LLM calls, we show that \method{} produces high-quality, data-driven, insightful feedback. Our analysis shows strong alignment with human-annotated errors and the ability to predict task success rate. 

Code: \url{https://ibm.biz/ACLEAR-Code}

\end{abstract}

\section{Introduction}

Agentic systems have become increasingly capable of defining strategies, executing actions, interacting with external environments, and solving complex, multi-step tasks~\cite{schick2023toolformer, Wang_2024}. 
This success has driven widespread adoption across various domains, including software engineering~\cite{Anthropic}, scientific discovery~\cite{ghafarollahi2025sciagents}, and open-ended web browsing~\cite{OpenAI}. Crucially, this paradigm shift is not limited to large-scale enterprise solutions. Individual developers are adopting agentic workflows to automate bespoke, day-to-day tasks. However, despite this democratization of agent building, agentic systems remain inherently brittle. They frequently exhibit subtle failure modes, repeated loops, misaligned sub-agent behavior, and error propagation across steps that are hard to detect from final outputs alone. 

This pressing need for oversight has led to the proliferation of agent observability platforms (e.g., \citeauthor{langsmith}, \citeauthor{langfuse}). While invaluable for logging execution traces, their evaluation capabilities are largely limited to basic metric aggregation or coarse, single-prompt LLM-as-a-judge assessments applied to the full trace. Consequently, developers are still required to manually inspect large numbers of traces to identify systemic issues. In parallel, the research community has focused on constructing agent error taxonomies~\cite{cemri2026why, zhu2026where, deshpande2025trail} and high-fidelity benchmarks~\cite{jimenez2024swebenchlanguagemodelsresolve, yehudai2025surveyevaluationllmbasedagents}. Yet, these approaches yield static, rigid categories or require extensive, hand-crafted engineering that cannot dynamically adapt to the bespoke tasks faced by everyday agent developers. 

In this work, to bridge this gap, we present \method{}, an automatic, dynamic, and easy-to-use evaluation method that produces rich, textual insights into agent behavior.
\method{} evaluates each trace, producing step-level and full-trace feedback, and then aggregates them across the full collection of execution traces to surface recurrent failures, quality degradation, and issues (See \S\ref{sec:method}). Our approach produces structured, textual diagnostics across three levels of granularity, the system, node, and trace levels, enabling developers to quickly understand not only \emph{what} failed, but \emph{why}.

\begin{figure*}[t]
    \centering
    \includegraphics[width=\textwidth]{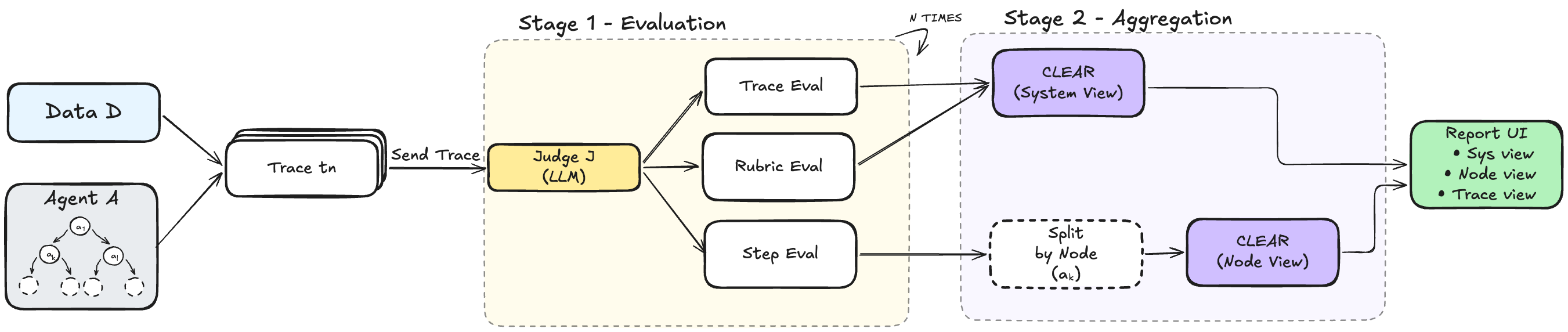}
    \caption{\method{} Pipeline. We start by preparing the execution traces. Stage 1: Apply multi-level per-trace evaluation via an LLM Judge. Stage 2: Aggregate insights using CLEAR, split into System-wide patterns and Node-specific patterns, and prepare them for the UI.}
    \label{fig:schema}
\end{figure*}

We provide \method{} as a pip-installable Python package designed for easy integration into existing agent development workflows (See \S\ref{sec:pipe}). It also provides an intuitive interactive UI for deep-dive trace analysis (See \S\ref{sec:ui}). Through experiments on diverse traces drawn from leading benchmarks and prominent agent architectures, we demonstrate that \method{} delivers actionable, high-level insights without requiring hand-crafted evaluation rubrics or extensive human annotation (See \S\ref{sec:results}). By lowering the barrier to meaningful agent evaluation and diagnostics, \method{} supports faster iteration, improved reliability, and more systematic understanding of agent behavior across tasks and domains.

In summary, our contributions are:
\begin{enumerate}
    \item \textbf{A Dynamic Evaluation Methodology:} We introduce a multi-level method that emphasizes automatic, dynamic, and granular evaluation insights.
    \item \textbf{Open-Source Package:} We provide a Python package with easy integration and an interactive visual dashboard.
    \item \textbf{Empirical Validation:} We demonstrate the efficacy of \method{} across varied benchmarks, agents, and models, showing its ability to surface execution failures without human-engineered tests. 
\end{enumerate}

We hope that \method{} will serve the broader NLP and software engineering communities, fostering faster iteration, improved agent reliability, and the development of next-generation evaluation tools.

\section{\method{} Method}\label{sec:method}
\method{} generates multi-level feedback by analyzing the agentic system behavior across an entire dataset. As described in Figure~\ref{fig:schema}, the pipeline ingests execution traces and outputs insights at the system, node, and trace levels.

Formally, let $\mathcal{D} = \{x_n\}_{n=1}^N$ be a dataset of $N$ tasks and $A$ be a target agentic system (i.e., a multi-agent system) composed of distinct nodes (e.g., sub-agents or components, depending on the development framework). Invoking $A$ on task $x_n$ yields an execution trace $t_n = \{(i_k, o_k, a_k)\}_{k=1}^{K_n}$, consisting of a sequence of LLM calls, where each call is divided into an input and an output pair, $(\{(i_k, o_k, a_k)\})$, produced by a specific node $a_k$, as dictated by the agent structure and execution flow. Overall, by running the agent on $\mathcal{D}$, we get the resulting traces, denoted as $\mathcal{T} = \{t_n\}_{n=1}^N$.

Given this data, our evaluation proceeds in two stages: trace evaluation and system-level aggregation.
As outlined in Algorithm~\ref{alg:method_pipeline}, first, for every trace $t_n$, we employ an LLM judge $J$ to perform three assessments:
(1) \textbf{Step-wise Evaluation:} For each pair $(i_k, o_k)$, $J_s$ produces a quality score and a natural language critique (subscript notation indicates different evaluation modes of $J$).
(2) \textbf{Trace-wise Evaluation:} Similarly, $J_t$ evaluates the quality of the complete trace, taking into account step and full trace considerations.
(3) \textbf{Rubric Evaluation:} We apply a two-step assessment. First, given $x_n$, the judge $J_r$ generates a set of task-specific criteria/rubrics required to accomplish the task. Then, based on $x_n$ and the generated rubrics $r_n$, the judge $J_v$ assesses whether these criteria were met within the trace $t_n$.

In the second stage, to identify high-level insights, we leverage CLEAR~\cite{yehudai2025clearerroranalysisllmasajudge} to cluster and summarize the instance-level feedback into global insights. For each node ($a_k$), we group input-output pairs associated with it, and apply CLEAR to surface component-specific failures ($\mathcal{I}_{node}$). Similarly, we aggregate trace-level judgments to identify holistic system behaviors ($\mathcal{I}_{sys}$). Finally, we also link each insight to the specific execution step or trace that triggered it.

This hierarchical approach delivers clear, interpretable insights across multiple levels of granularity, giving the agent developer visibility into the system at different resolutions, from fine‑grained nodes and traces to the full system view.

\begin{algorithm*}[t]
\fontsize{11pt}{11.5pt}\selectfont % <--- CUSTOM SIZE BETWEEN REGULAR AND SMALL

\caption{\method{} Insight Generation Pipeline}
\label{alg:method_pipeline}

% --- APPLY THE COMMENT STYLE HERE ---
\SetCommentSty{mycommfont}
% ------------------------------------

\DontPrintSemicolon
\SetKwInput{Input}{Input}
\SetKwInput{Output}{Output}

\Input{Dataset $\mathcal{D} = \{x_n\}_{n=1}^N$; Agent $A$; Judge $J$; Aggregator \textsc{Clear}}
\Output{System Insights $\mathcal{I}_{sys}$, Node Insights $\mathcal{I}_{node}$, Trace Evaluations $\mathcal{E}_{trace}$}

% Initialize containers
$\Phi_{node} \leftarrow \emptyset$; $\Phi_{sys} \leftarrow \emptyset$ \tcp*{Init feedback containers}

\tcc{Stage 1: Execution \& Granular Evaluation}
\For{$n \leftarrow 1$ \KwTo $N$}{
    $t_n \leftarrow \text{Execute}(A, x_n)$ \tcp*{Trace $t_n = \{(i_k, o_k, a_k)\}$ with inputs, outputs, nodes}
    
    \tcp{1. Node-wise Evaluation}
    \ForEach{step $k$ in $t_n$}{
        $c_{n_k}^{node} \leftarrow J_s(i_k, o_k)$ \tcp*{Critique individual step}
        $\Phi_{node}[a_k] \leftarrow \Phi_{node}[a_k] \cup \{c_{n_k}^{node}\}$ \tcp*{Group by node $a_k$}
    }
    
    \tcp{2. Trace-wise Evaluation}
    $c^{trace}_n \leftarrow J_t(t_n)$ \tcp*{Holistic trace critique}
    
    \tcp{3. Rubric Evaluation}
    $r_n \leftarrow J_r(x_n)$ \tcp*{Generate task-specific criteria}
    $c^{rubric}_n \leftarrow J_v(t_n, r_n)$ \tcp*{Check compliance}
    
    $\Phi_{sys} \leftarrow \Phi_{sys} \cup \{c^{trace}_n, c^{rubric}_n\}$ \tcp*{Collect for system view}
    $\mathcal{E}_{trace}[n] \leftarrow \{c^{trace}_n, c^{rubric}_n, c_{n_k}^{node}\}$ 
}

\tcc{Stage 2: Insight Aggregation via CLEAR}
$\mathcal{I}_{node} \leftarrow \emptyset$\;
\ForEach{node $a$ in $\Phi_{node}.\text{keys}$}{
    $\mathcal{I}_{node}[a] \leftarrow \textsc{Clear}(\Phi_{node}[a])$ \tcp*{Per-node recurring patterns}
}

$\mathcal{I}_{sys} \leftarrow \textsc{Clear}(\Phi_{sys})$ \tcp*{Global system patterns}

\Return $\mathcal{I}_{sys}, \mathcal{I}_{node}, \mathcal{E}_{trace}$
\end{algorithm*}

% backend - pipeline
\section{\method{} Framework}\label{sec:framwork}

\subsection{Pipeline}\label{sec:pipe}

To allow easy integration and usability, we provide \method{} as a Python package available on PyPI (Permissive Apache 2.0 License). The package supports the different end-to-end evaluation levels described in \S\ref{sec:method}.
Each evaluation level in the pipeline can be used on its own or combined with the others, allowing users to tailor the workflow to their specific evaluation needs and preferences.

For easy onboarding, we adopt an OpenTelemetry\footnote{\href{https://opentelemetry.io/}{OpenTelemetry}
}-compatible format. Specifically, we utilize LangFuse-formatted\footnote{\href{https://langfuse.com/integrations/native/opentelemetry}{LangFuse}} traces, which we convert to an intermediate representation that serves as input to the pipeline. For other trace formats, we require only minimal preprocessing to reach the same intermediate state that captures the LLM call's inputs and outputs in the trace, along with the necessary metadata. We focus our analysis on the LLM interactions, as they govern the system's decision-making and are its most stochastic element.

We design specific prompts for each judge evaluation mode. For $J_s$, the judge assesses step‑level aspects such as correctness, completeness, and clarity. For $J_t$, we extend these criteria to trace‑level dimensions, including execution quality and the final deliverable. In $J_r$, the judge needs to decide on the number of rubrics and generate them to suit the given task. Each prompt elicits a brief textual justification prior to the score, functioning as a chain-of-thought rationale. While our method primarily focuses on providing textual insights, we also surface these quantitative scores in the UI. When ground‑truth evaluation scores are available for each trace, the system generates further insights into execution paths and predictive patterns of trace success, and additionally assesses the reliability of the judge. All the prompts are presented in App.~\ref{app:prompts}.
To support customization, users can adjust the evaluation dimensions, override the prompts, or replace the judge with a custom Python implementation. 

\begin{figure*}[ht!]
    \begin{center}        
    % Trim format: trim={left bottom right top}
    % Here we are trimming 0 from left, 150pt from bottom, 0 from right, 0 from top.
    \includegraphics[
        width=1\linewidth, 
        trim={0 100pt 0 0}, 
       clip
    ]{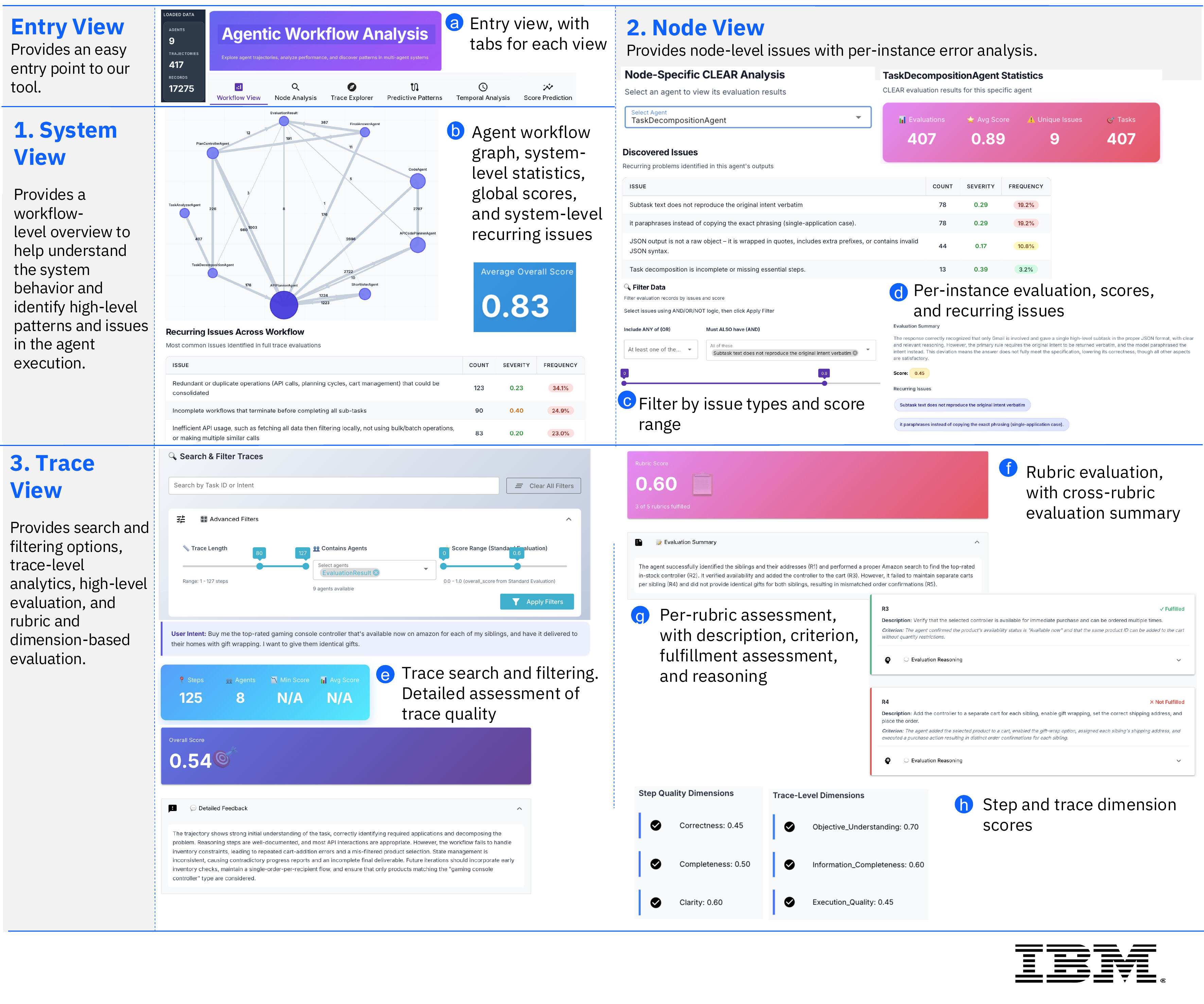} 
    \caption{
    The interactive UI of \method{}, enabling multi-granular evaluation and diagnosis of agentic workflows. (a) The entry module provides top-level navigation across different analysis tabs. (b) The \textbf{System View} offers a macro-level summary, visualizing agent topologies, global performance scores, and system-wide recurring issues. The \textbf{Node View} facilitates agent-specific error analysis via (c) issue- and score-based filtering to isolate relevant evaluations, alongside (d) per-instance scoring and error distributions. The \textbf{Trace View} enables fine-grained, instance-level inspection, featuring (e) trace search and filtering capabilities, (f) cross-rubric evaluation summaries, (g) detailed per-rubric assessments with fulfillment reasoning, and (h) granular step- and trace-level dimension scores.}
    \label{fig:ui}
    \end{center}
\end{figure*}

\paragraph{Code}
We provide \method{} as a PyPI package. The analysis can be executed with a single CLI command, configured via a YAML file. Once processing completes, the interactive interface can be launched from the command line.
The pipeline stores its results as a ZIP file in the designated output directory, which can then be loaded manually into the app.

\begin{myshellbox}
\begin{lstlisting}
 $ pip install clear-eval
\end{lstlisting}
\end{myshellbox}

% fornend - UI
\subsection{\method{} UI}\label{sec:ui}

\method{} dashboard (Figure~\ref{fig:ui}) provides a hierarchical visual suite. We designed it to move beyond static telemetry, enabling agent developers and researchers to diagnose agent behaviors across levels.
The interface is structured around three primary perspectives:

\paragraph{System Level} This view dynamically reconstructs the multi-agent topology directly from execution traces. It presents high-level agent behavioral patterns, like node usage and flow dynamics. Finally, it aggregates global performance scores and surfaces systemic recurring issues.

\paragraph{Node View} This view allows navigating between agent nodes. For each, it presents the dynamically generated issues the node exhibits. Users can filter steps by issue types and score ranges. This allows targeted inspection of per-instance error distributions, surfacing recurring patterns localized to individual prompts or behaviors.

\paragraph{Trace View} Facilitating fine-grained analysis, the Trace View unpacks individual execution traces. It presents overall trace evaluation, alongside granular, step-level dimension scores, and rubric evaluation. Crucially, it exposes the LLM judge's natural language reasoning for each assessment, providing users with interpretable, context-aware justifications for every identified failure mode.

\section{Experimental Setup}
% \label{sec:case}

% \subsection{Experimental Setup}
To rigorously evaluate \method{} across diverse settings, we curate execution traces generated by leading agent architectures and LLMs across prominent benchmarks. 
Specifically, we take traces from the following benchmarks: \SWEBench~\cite{jimenez2024swebenchlanguagemodelsresolve}, \GAIA~\cite{mialon2023gaiabenchmarkgeneralai}, \AppWorld~\cite{trivedi2024appworldcontrollableworldapps}, and \Tbench~\cite{barres2025tau2benchevaluatingconversationalagents}. The agents are CUGA~\cite{marreed2025enterprisereadycomputerusinggeneralist}, the SOTA agent on AppWorld, HAL generalist agent~\cite{kapoor2026holistic}, and Hugging Face's Open Deep Research agent~\cite{roucher2025opendeepresearch}, with top OpenAI and Anthropic models (See Table \ref{tab:datasets}).

We collect traces from HAL~\cite{kapoor2026holistic}, TRAIL~\cite{deshpande2025trail}, and the AppWorld leaderboard, and consolidate them into our unified intermediate representation schema. We select seven settings to support comparative analyses across models, agents, and benchmarks. We present detailed descriptions of the benchmarks, the evaluated agents, and the specific trace datasets in Appendix \ref{app:data_details}.

\begin{table}[t]
    \centering
    \resizebox{\columnwidth}{!}{%
        \renewcommand{\arraystretch}{1.2}
        \begin{tabular}{@{}lllcl@{}}
            \toprule
            \textbf{Benchmark} & \textbf{Agent} & \textbf{Model} & \textbf{\# Traces} & \textbf{Source} \\
            \midrule
            AppWorld & CUGA & GPT-4o & 417 & Leaderboard\\
            \midrule
            GAIA & Generalist Agent & Claude 4.5 Sonnet & 165 & HAL \\
                 & Generalist Agent & GPT-4.1 & 165 & HAL \\
                 & HF DeepResearch & Claude 4.5 Sonnet & 165 & HAL \\
                 & HF DeepResearch & OpenAI o3 & 117 & TRAIL \\
            \midrule
            SWE-bench Ver. Mini & Generalist Agent & Claude 4.5 Sonnet & 50 & HAL\\
            \midrule
            TAU-bench (Airline) & Generalist Agent & Claude 3.7 Sonnet & 50 & HAL\\
            \bottomrule
        \end{tabular}%
    }
    \caption{Data statistics for the curated traces}
    \label{tab:datasets}
\end{table}

As judges, we employ two leading models, 
OSS-120B~\cite{openai2025gptoss120bgptoss20bmodel} in high thinking mode as a representative of a leading open-source model, and GPT-5~\cite{singh2025openaigpt5card} as a closed-source model.

We perform trace-wise evaluation across all seven trace datasets using two judge models. The resulting evaluations are then passed to the CLEAR aggregation stage for issue discovery.

\section{\method{} Issues Results}\label{sec:results}

In the following, we report findings on the universal failure patterns, the effect of the agent architecture and the backbone model, benchmark-specific issues, and the impact of judge selection.

\paragraph{Universal Error Patterns}
Several recurring issue categories appeared among the 195 trace-level issues generated across all configurations, reflecting systemic weaknesses in current agent systems: (1) Redundant and Inefficient Tool Usage: unnecessary repeated calls, poorly designed queries, or wasted computation; (2) Insufficient Error Handling and Recovery: agents frequently failed to recover from tool errors or to shift to alternative strategies after failure and lacked effective fallback mechanisms; (3) Incomplete Workflows: agents failed to bring tasks to completion and fulfill all goals; (4) 
Output Formatting and Schema Compliance: agents failed to adhere to output formats.

\paragraph{Domain-Specific Issues}
(a) System-Level: Beyond these shared errors, each benchmark displayed its own domain‑specific weaknesses. \GAIA, a research-oriented benchmark, was dominated by \textbf{sourcing and verification failures} (e.g., \textit{``Lack of cross-verification across independent sources''}); 
AppWorld, which tests multi‑step API orchestration, exhibited unique failures such as \textbf{incomplete executions and domain‑specific workflow breakdowns} (e.g.,
\textit{``acting on contaminated shopping carts and dropping email attachments''});
Results on \SWEBench highlight code-related issues, such as \textbf{monkey-patching and broken diff output},  while \Tbench focused on \textbf{policy violations} (e.g., \textit{``unauthorized payment selection, fabricated cost estimates''}). 
Notably, \method{} discovered these domain-specific issues without any benchmark-specific prompting. 
\newline(b) Node-level: This differentiation extends further at the node level. Running our method on the CUGA agent reveals that while universal issues like \textbf{JSON malformation} appeared across nearly all nodes, different nodes surfaced distinct failure types matching their role: planning nodes were dominated by \textbf{task decomposition} and \textbf{API selection} issues (e.g., TaskDecompositionAgent: \textit{``subtasks are ordered illogically or not in a natural execution sequence''}), while execution nodes surfaced functional bugs (e.g., APICodePlannerAgent: \textit{``missing pagination handling for APIs that return multiple pages of results''}). Moreover, this evaluation mode allows pinpointing specific pitfalls behind each failure mode and addressing them directly. For example, hallucinations occur mainly during the planning stages (e.g., ShortlisterAgent: \textit{``APIs not defined in the supplied API catalog are listed''}) but not during execution. Insights like these help agent developers fine‑tune the relevant 
components more effectively. See Appendix \ref{app:examples} for concrete examples of both cross-benchmark and cross-level issue variations.

\paragraph{Backbone Model and Agent Differences}
Comparing GPT-4.1 and Claude 4.5 Sonnet as backbones for the HAL agent on GAIA (judged by GPT-5), the two models shared the majority of their system-level failure profile: both were flagged for \textbf{source verification gaps, tool misuse, and output formatting noncompliance}. For instance, both produced nearly identical issues around output compliance (GPT-4.1: \textit{``noncompliance with required execution and output formats/protocols''}; Claude 4.5 Sonnet: \textit{``failure to adhere to output formatting and deliverable specifications''}). However, each also exhibited unique tendencies: GPT-4.1 was flagged for \textit{``prematurely giving up after errors instead of diagnosing, retrying, or pivoting to alternatives''}, while Claude 4.5 Sonnet was associated with \textit{``contradictory or self-conflicting statements; does not commit to a consistent interpretation''}.
Similarly, comparing the HF DeepResearch and HAL agents with Claude as the backbone over \GAIA reveals a largely shared error profile, with some small distinctions, suggesting the dataset has a greater effect than the agent architecture on the error types. 

\paragraph{Judge Selection} Both judges were consistently able to uncover diverse and non‑trivial recurring issues. However, they produced qualitatively different diagnoses, even of the same agent behavior. Their output differed not only in wording but also in depth, specificity, and the behavior they chose to emphasize. OSS-120B tended to generate shorter issues (67 vs. 130 characters on average) and to surface broader and more generic categories, more focused on operationally oriented failures (e.g., ``Redundant searches and file inspections causing inefficiency'' or ``Misused tool arguments or invoked the wrong tool'' on \SWEBench). In contrast, GPT-5 produced longer, more nuanced, and domain-specific failure modes that more frequently targeted verification and validation failures, incorrect logic or reasoning, and methodological correctness (e.g., ``breaks SQL query correctness due to missing alias remapping when combining SQL components''). 
These findings suggest that judge selection is consequential for determining the specificity and depth of the generated failures.
% design decision

\input{latex/tables/trail_alignment}

\input{latex/tables/score_prediction}
\section{Analysis}
% To verify our method, we conduct two analyses.
We validate \method{} through two complementary analyses.
The first compares our issues against human-annotated errors. The second compares our score prediction methods with a few ground-truth benchmarks' labels.
 
\subsection{Alignment with Human Error Taxonomies}
% \section{Alignment with Human Taxonomy}\label{app:trail_exp}
To validate that our automatically generated issues capture meaningful error patterns, we first perform a semantic mapping between our generated issues and \Trail categories~\cite{deshpande2025trail}. \Trail provides a hierarchical taxonomy of 20 error categories spanning reasoning, planning, and system execution failures. Here, we use the 12 non-execution categories as \method{} focuses on LLM reasoning and planning. These categories account for 94\% of the ground-truth labels.

Since our issues are taxonomy-free by design, we first apply a semantic alignment: we map each of our system-level issues into the \Trail categories as either a full match (directly corresponding to a \Trail category), or a partial match (overlaps conceptually but covers a broader or adjacent concern). The full mappings between the issues produced by both judges and the \Trail taxonomy are presented in Appendix \ref{app:trail_map}.

The mapping was performed using Claude Opus 4.6 and verified by the authors. All 15 GPT-5 issues and all 12 OSS-120B issues map to at least one TRAIL category, collectively covering 
% all 12 and 10 of the 12 relevant categories, respectively.
12 and 10 of the 12 relevant categories, respectively.
% The full GPT-5 and OSS-120b mappings are presented in Appendix~\ref{app:mapping}.

To verify that the alignment holds at the instance level, i.e., traces flagged with issues by \method{} exhibit the corresponding \Trail errors, we propagate the mapping transitively to individual traces (117 in total) and measure agreement. We report macro-averaged F1 as the primary metric, as it equally weights all error categories and thus directly measures breadth of taxonomy coverage. To calibrate, we compare against two baselines: a random predictor weighted by the true category frequencies, and a majority baseline that always predicts the four most common categories. 

Table~\ref{tab:error_cat} presents the results. The GPT-5 judge achieves the strongest agreement under the full+partial matching, with a macro-F1 of 0.459 and micro-F1 of 0.497. The frequency baseline is competitive on micro-F1 (0.459) due to the skewed category distribution, but its low macro-F1 (0.199) indicates that it fails to cover the tail of the error distribution. As expected, the GPT-5 judge outperforms the smaller OSS-120B judge.

Overall, \method{} recovers the majority of reasoning and planning error categories without requiring predefined category definitions. The generated issues are often more fine-grained and actionable than the TRAIL categories they map to, capturing specific failure patterns where the taxonomy provides only broad groupings. This suggests that our method can preserve the diagnostic capabilities of expert taxonomies while surfacing more targeted and nuanced insights. 

\subsection{Score Prediction}
To evaluate our judge's ability to predict trace success, we compute the area under the ROC curve (AUC) between the ground-truth and the predicted scores. \method{} provides three methods to predict trace success: (1) \textbf{Trace}: the overall score generated by the trace-wise evaluation; (2) \textbf{Rubric}: the proportion of task-level rubrics predicted as fulfilled; and (3) \textbf{Step-wise}: the average score across all steps within the trace.

Table \ref{tab:score_pred} presents the full results. GPT-5 generally outperforms OSS-120B across the methods. 
Across configurations, the trace-level method is the strongest predictor, outperforming the step-wise and rubric methods.
% This likely reflects the different assumptions of each method that do not always hold. 
This likely reflects the fact that the underlying assumptions of each method do not hold uniformly across all settings.
The rubric method assumes the task description contains all the requirements to determine success, which breaks for \Tbench when implicit policy adherence is critical. The step-wise method assumes that trace evaluation can be decomposed into the quality of isolated steps, which is more suitable for some agents and benchmarks. For instance, modular agent architectures like CUGA are composed of distinct, self-contained components with clearly defined tasks, making it easier to assess each node's contribution. It can also benefit from a composable task structure, like in \SWEBench, where the tasks naturally decompose into discrete phases, e.g., locating, understanding, and fixing a bug.
% These decomposability qualities make step-level quality tracks overall success.
These differences suggest that \method{} evaluation methods can provide complementary signals depending on the target domain and agent.

Comparing results across benchmarks reveals large variations.
\AppWorld is the most predictable benchmark, with all results exceeding 0.75, and GPT-5 specifically achieving at least 0.82 AUC with all methods. \Tbench results, on the other hand, do not exceed 0.62, and both \GAIA and \SWEBench exhibit variation depending on the method, agent, and model.
These results call for further research to investigate the effectiveness of trace judges in different agentic configurations.

\subsection{Rubric Analysis}
To better understand what our generated rubrics capture, we compared our generated rubrics against benchmark-native evaluation criteria on sampled tasks. Notably, our rubrics are generated from the task description alone, without access to benchmark-internal metadata. This gap affects benchmarks differently.
In \AppWorld, our rubrics correctly capture the expected agent behavior, but describe the process qualitatively, while gold assertions are programmatic state checks against precomputed outcomes. For example, on a shopping task, our rubrics capture the workflow (retrieve list, parse items, add to cart, checkout), while gold constraints focus on assertions that validate the gold state is reached, like "exactly one new order created" and "no address records modified.
In \Tbench, the difference is much starker. Many tasks are adversarial, meaning the correct behavior is to refuse the user's request. Because our generator sees only the surface-level request, it inadvertently produces rubrics that reward task completion instead.
These findings suggest that rubric generation from task descriptions alone can be enhanced by task metadata and should be examined based on the target agentic setting.

\section{Related Work}

\paragraph{General Agent Evaluation}
Our work takes a first step towards automatic environment-agnostic agent evaluation.
Recent work has begun to understand the importance of standardizing agentic evaluation~\citep{bandel2026agentic, bandel2026readyforgeneral} and has made first steps towards achieving it~\citep{kapoor2026holistic}. These efforts focus on the runtime and execution layers across environment type~\citep{bandel2026generalagentevaluation, Harbor_Framework}, standardizing agent evaluation protocols~\citep{lacoste2026cubestandardunifyingagent}, and building frameworks that enable easy and scalable agent and benchmark integration.
While these efforts focus on standardizing the benchmarking infrastructure, \method{} operates above the execution layer and addresses how to interpret traces, providing out-of-the-box multi-level agent evaluation.

\paragraph{Agent Meta-Evaluation}
Several recent works focus on creating benchmarks that assess judges' ability to detect agent erroneous steps and classify them into the right pre-defined category~\cite{cemri2026why, zhu2026where, deshpande2025trail, lu2025agentrewardbench}.
These works extend a large body of works on meta-evaluation of LLMs~\cite{zheng2023llmaaj, gera2025justrankbenchmarkingllmjudges}.
Unlike these approaches, which assume a fixed error taxonomy and evaluate judges' recovery of it, \method{} operates without predefined categories, dynamically surfacing failure patterns that adapt to the target system and domain.

\section{Conclusions}
We presented Agentic CLEAR, an automatic evaluation framework that produces multi-level textual insights into agent behavior at scale, without requiring predefined error taxonomies or hand-crafted rubrics. Across four benchmarks and seven agentic configurations, we demonstrated alignment with human-annotated errors and meaningful predictive signal for task success. Key directions for future work include extending \method{} to analyze system execution alongside reasoning and planning, improving judge capabilities and reliability across diverse agentic settings, and enabling systematic cross-configuration comparisons.

\bibliography{custom}

\appendix
\input{appendix}

\end{document}

%% file: commands.tex
\usepackage{caption}
\usepackage[export]{adjustbox}

\usepackage{subfig}

\usepackage{booktabs}
\usepackage{hyperref}
\usepackage{longtable}

\usepackage{bbm}
\usepackage{stmaryrd}
\usepackage{amsmath,amssymb}

\usepackage{amsthm}

\usepackage{refcount}
\usepackage{fancyvrb}

\newcommand{\method}{Agentic CLEAR}
\newcommand{\Tbench}{$\tau^{2}$-Bench\xspace}
\newcommand{\AppWorld}{AppWorld\xspace}
\newcommand{\SWEBench}{SWE-Bench Verified Mini\xspace}
\newcommand{\GAIA}{GAIA\xspace}

\newcommand{\HAL}{HAL\xspace}
\newcommand{\Trail}{TRAIL\xspace}

\usepackage{fix-cm}
\usepackage{xcolor}

\definecolor{algteal}{HTML}{008080}

\usepackage{booktabs}
\usepackage{multirow}
\usepackage{hyperref}
\usepackage{listings}
\usepackage[most]{tcolorbox}
\usepackage{xcolor}

\definecolor{myred}{HTML}{EC6460}   % Replace FF0000 with your desired HTML code for red
\definecolor{myyellow}{HTML}{ECBD60  } % Replace FFFF00 with your desired HTML code for yellow
\definecolor{mygreen}{HTML}{90EC60}  % Replace 008000 with your desired HTML code for green
\definecolor{mygray}{HTML}{AEB0AC}   % Replace 808080 with your desired HTML code for gray#3c3273
\definecolor{deeppurple}{HTML}{3c3273}
\definecolor{mygray}{RGB}{240,240,240}
\definecolor{darkgray}{RGB}{64,64,64} % This is a dark gray. Adjust the numbers to change the shade.

\newtcolorbox{mycodebox}{
    enhanced,
    fonttitle=\bfseries,
    colback=white,
    colframe=mygray,
    colbacktitle=mygray,
    coltitle=black,
    rounded corners,
    boxrule=0.5mm,
    top=5mm,
    interior style={top color=mygray, bottom color=white},
    overlay={
        \node[anchor=west, font=\bfseries] at ([xshift=5pt, yshift=-10pt]frame.north west)
        {\textcolor{myred}{$\bullet$} \textcolor{myyellow}{$\bullet$} \textcolor{mygreen}{$\bullet$}};
    },
    left=4mm,
    right=2mm,
    bottom=2mm,
}

\newtcolorbox{myshellbox}{
    enhanced,
    fonttitle=\bfseries,
    colback=white,
    colframe=mygray,
    coltext=white,
    colbacktitle=mygray,
    coltitle=black,
    rounded corners,
    boxrule=0.5mm,
    top=5mm,
    interior style={top color=black, bottom color=darkgray},
    overlay={
        \node[anchor=west, font=\bfseries] at ([xshift=5pt, yshift=-10pt]frame.north west)
        {\textcolor{myred}{$\bullet$} \textcolor{myyellow}{$\bullet$} \textcolor{mygreen}{$\bullet$}};
    },
    left=2mm,
    right=2mm,
    bottom=2mm,
}

\usepackage[utf8]{inputenc}
\usepackage{tcolorbox}
\usepackage{fvextra}        % Enables \VerbatimInput with breaklines
\usepackage{caption}
\usepackage{graphicx} 
\usepackage{dashrule}

%% file: latex/tables/trail_alignment.tex
\begin{table}[t]
\centering
\small
\begin{tabular}{lcc}
\hline
\textbf{Method} & \textbf{Micro F1} & \textbf{Macro Cat F1} \\
\hline
Random (GT freq)            & 0.342 & 0.288 \\
Always top-4                & 0.459 & 0.199 \\
OSS-120B (full)             & 0.377 & 0.261 \\
OSS-120B (full+partial)     & 0.427 & 0.374 \\
GPT-5 (full)                & 0.467 & 0.368 \\
\textbf{GPT-5 (full+partial)} & \textbf{0.497} & \textbf{0.459} \\
\hline
\end{tabular}
\caption{Error category prediction performance against TRAIL (Planning and Reasoning categories).}
\label{tab:error_cat}
\end{table}

%% file: latex/tables/score_prediction.tex
\begin{table*}[t]
\centering
\resizebox{\textwidth}{!}{%
\setlength{\tabcolsep}{5pt} % adjust spacing between columns
\begin{tabular}{@{}lllcccccc@{}}
\toprule
 & & & \multicolumn{2}{c}{\textbf{Step-Wise}} & \multicolumn{2}{c}{\textbf{Trace}} & \multicolumn{2}{c}{\textbf{Rubric}} \\
\cmidrule(lr){4-5} \cmidrule(lr){6-7} \cmidrule(lr){8-9}
\textbf{Benchmark} & \textbf{Agent} & \textbf{Model} & \textbf{OSS-120B} & \textbf{GPT-5} & \textbf{OSS-120B} & \textbf{GPT-5} & \textbf{OSS-120B} & \textbf{GPT-5} \\
\midrule
AppWorld & CUGA & GPT-4o & 0.758 & 0.823 & 0.837 & 0.890 & 0.778 & 0.828 \\
\midrule
GAIA & Generalist Agent & Claude 4.5 Sonnet & 0.664 & 0.632 & 0.712 & 0.720 & 0.596 & 0.566 \\
GAIA & Generalist Agent & GPT-4.1 & 0.707 & 0.742 & 0.839 & 0.848 & 0.560 & 0.597 \\
GAIA & HF DeepResearch & Claude 4.5 Sonnet & 0.541 & 0.546 & 0.609 & 0.716 & 0.537 & 0.571 \\
GAIA & HF DeepResearch & OpenAI o3 & 0.783 & 0.706 & 0.774 & 0.819 & 0.729 & 0.736 \\
\midrule
SWE-bench & Generalist Agent & Claude 4.5 Sonnet & 0.788 & 0.779 & 0.661 & 0.803 & 0.505 & 0.524 \\
\midrule
TAU-bench & Generalist Agent & Claude 3.7 Sonnet & 0.409 & 0.529 & 0.618 & 0.554 & 0.539 & 0.597 \\
\bottomrule
\end{tabular}%
}
\caption{AUC for predicting trajectory success using Agentic CLEAR scores. We report trace-level, rubric-based, and step-wise (average) scores}
\label{tab:score_pred}
% \vspace{-10pt}
\end{table*}

%% file: appendix.tex
\newpage
\input{latex/tables/issue_examples}
\input{latex/tables/issues_cuga}

\section{Prompts}\label{app:prompts}
The prompts are presented in our \href{https://github.com/IBM/CLEAR}{code repository}.

\section{Data}\label{app:data_details}

\subsection{Benchmarks}

\paragraph{\SWEBench} A subset of 50 human-validated real-world software engineering tasks from popular Python repositories. Each provides a GitHub issue and repository snapshot; agents produce patches that are evaluated against hidden unit tests~\cite{jimenez2024swebenchlanguagemodelsresolve}.

\paragraph{\AppWorld} A benchmark for evaluating user-assistance agents on realistic day-to-day digital tasks. 
The agent interacts with the environment by writing Python code that is executed in a dedicated interpreter with access to the \AppWorld APIs~\cite{trivedi2024appworldcontrollableworldapps}. 

\paragraph{\Tbench} evaluates customer-service agents across retail, airline, and telecom domains via LLM-simulated users, measuring both policy-compliant task completion and violation rejection~\cite{barres2025tau2benchevaluatingconversationalagents}.

\paragraph{\GAIA} comprises 466 human-designed, real-world questions for evaluating general AI assistants. Each task requires fundamental abilities such as web browsing, multi-modal, and multi-file handling~\cite{mialon2023gaiabenchmarkgeneralai}.
% \mss{for tau or GAIA do we have multiple agents on the same benchmark? if so, we could have shown different issues for different agent which could make it interesting}

\subsection{Traces Data}

\paragraph{\HAL} A unified evaluation framework that standardizes agent benchmarking across diverse domains. It provides a large set of execution traces, enabling automated trace evaluation to uncover hidden failure modes, issues in agent behavior, and unsafe real-world actions~\cite{kapoor2026holistic}.

\paragraph{\Trail} provides a set of execution traces with human-annotated agent errors, based on a predefined taxonomy,
testing whether LLM judges can accurately pinpoint reasoning, planning, and system execution failures~\cite{deshpande2025trail}.

\subsection{Agents}

\paragraph{CUGA}
\textbf{C}onfig\textbf{U}rable \textbf{G}eneralist \textbf{A}gent (CUGA) is an open-source system specifically designed for enterprise automation. It handles complex tasks through multi-agent orchestration, dynamic reasoning, and API integrations while ensuring strict policy compliance.

\paragraph{HAL Generalist Agent} An agent developed by the HAL team, designed to work across their unified evaluation framework.

\paragraph{\textbf{HF Open Deep Research Agent}} An open-source agentic search framework developed by Hugging Face. It is engineered to autonomously navigate the web, synthesize information across long trajectories, and generate comprehensive, citation-backed answers for complex research queries.

\section {Issue Examples}\label{app:examples}
We present two examples illustrating the issues generated by \method{} across different configurations and analysis levels.

Table \ref{tab:issues_benchmark} presents the top 10 system-level issues discovered for both \GAIA and \SWEBench under the same configuration (same agent, model, and judge). Four of the issues are shared across the benchmarks, capturing universal error patterns such as inefficient workflows or tool misuse. The remaining issues are domain-specific: \GAIA surfaces issues like inadequate source verification or unreliable data processing, while \SWEBench reveals engineering-oriented failures such as broken patch output or missing regression tests. This differentiation occurred without any benchmark-specific prompting, demonstrating \method{}'s ability to adapt issue discovery to the relevant data.

Table \ref{tab:issues_system_node} presents the top 10 issues discovered at the system level and at the node level for the TaskDecompositionAgent, both generated from the same CUGA traces on AppWorld. The system-level issues capture system-wide failure modes such as incomplete task execution or entity resolution errors. The node-level issues pinpoint planning-stage errors, including wrong app assignments or unsupported capability assumptions. Several themes appear at both levels but with different granularity. For example, the system level flags incomplete execution, while the node level traces it to the TaskDecompositionAgent omitting the finalization step. Together, the two views offer complementary diagnostics: the system level surfaces broad failure patterns, while the node level localizes problems to specific components and uncovers nuanced failures not visible at the system level.

\input{latex/tables/trail_map_gpt5}
\input{latex/tables/trail_map_120b}

\section {\method{} Issues to \Trail Mapping}\label{app:trail_map}
Tables \ref{tab:gpt5-trail-mapping} and \ref{tab:oss-trail-mapping} present the full mapping between the issues generated by \method{} at the system-level using GPT-5 and OSS-120B, respectively, and the \Trail taxonomy.

%% file: latex/tables/issue_examples.tex
\begin{table*}[t]
\centering
\small
\begin{tabular}{p{0.46\textwidth} p{0.46\textwidth}}
\toprule
\textbf{GAIA} & \textbf{SWE-bench Verified Mini} \\
\midrule
\multicolumn{2}{c}{\textit{Similar Issues}} \\
\midrule
\addlinespace
Inefficient workflow that delays or neglects high-signal resources (e.g., provided local files/attachments), causing redundant searches and retries
& Employs inefficient, noisy workflows with unnecessary detours and retries \\
\addlinespace
Violates tool constraints or misuses provided tools (e.g., disallowed \texttt{open}/\texttt{import}/\texttt{apt-get})
& Misuses tools or output formatting (disallowed imports, use of \texttt{open}/\texttt{subprocess}, incorrect code fences) \\
\addlinespace
Failure to adhere to output formatting and deliverable specifications (e.g., missing required terminators, wrong format)
& Does not follow task instructions or required outputs (e.g., missing facts survey, plan, or final patch) \\
\addlinespace
Inadequate edge-case handling in computations (e.g., zero derivative, convergence and rounding rules)
& Delivers partial fixes that miss edge cases or cross-backend/platform differences \\
\midrule
\multicolumn{2}{c}{\textit{Benchmark-Specific Issues}} \\
\midrule
\addlinespace
Failure to use and verify the mandated authoritative source and its exact version/timeframe; reliance on mirrors/snippets
& Produces incorrect or non-applicable patch output (escaped content, partial diffs, missing unified diff) \\
\addlinespace
Insufficient cross-validation and evidentiary support; claims presented without corroboration
& Runs commands/tests without ensuring environment prerequisites and paths are correct \\
\addlinespace
Unreliable data processing: fragile parsing and incorrect filtering logic
& Introduces broad behavior changes without proper scoping or compatibility/regression analysis \\
\addlinespace
Wrong methodological framework or inconsistent formalism for the task
& Avoids established APIs/patterns and relies on fragile techniques like monkey-patching \\
\addlinespace
Incomplete enumeration or coverage before counting (missing items/pages; partial lists)
& Insufficient validation of changes (lacks repo tests/regression, skips context-specific checks) \\
\addlinespace
Poor disambiguation of task terms or scope, leading to misinterpretation of requirements
& Provides no comments or documentation explaining rationale and potential impacts \\
\bottomrule
\end{tabular}
\caption{System-level issues generated by Agentic CLEAR for two benchmarks using the same agent (HAL Generalist), model (Claude 4.5 Sonnet), and judge (GPT-5). Top: shared issues surfaced for both benchmarks. Bottom: benchmark-specific issues.}
\label{tab:issues_benchmark}
\end{table*}

% \begin{table}[h]
% \small
% \centering
% \caption{}
% \label{tab:issues_judges}
% \resizebox{0.5\textwidth}{!}{%
% \begin{tabular}{@{}lccc@{}}
% \toprule
% \textbf{} \\
% \midrule

% \bottomrule
% \end{tabular}
% }
% \end{table}

%% file: latex/tables/issues_cuga.tex
\begin{table*}[t]
\centering
\small
\begin{tabular}{p{0.46\textwidth} p{0.46\textwidth}}
\toprule
\textbf{System-Level Issues} & \textbf{TaskDecompositionAgent (Node-Level)} \\
\midrule
\addlinespace
Execution flow management and processing strategy flaws: inefficient execution and incomplete coverage, loss of critical variables across steps, and continuing after success
& Assumes unsupported app capabilities; violates strict app constraints \\
\addlinespace
Validation and preconditions gaps: skips critical pre/post-action checks, neglects idempotency, fails to pre-check required permissions/tokens, and claims success without evidence
& Workflow coherence errors: reasoning, tasks, and ordering don't align; steps lack clear dependencies or references to prior outputs \\
\addlinespace
Blockage handling failures: does not prompt for missing info, avoids practical fallbacks or alternative paths, and declares failure prematurely
& Fails to return the user's intent verbatim for single-app tasks; paraphrases or alters details \\
\addlinespace
Incomplete execution: stops short of finishing the core task (checkout, send/forward, save/receipt)
& App boundary and handoff mistakes: wrong app assignment, unclear division of responsibilities, or cross-app access without explicit handoff \\
\addlinespace
Entity resolution weaknesses: brittle matching/normalization and poor disambiguation among multiple candidates
& Insufficient disambiguation criteria for selecting among multiple emails/items or handling time zones \\
\addlinespace
API/framework contract noncompliance: schema/role/output violations, misdeclared capabilities, misuse of fields/params, or using interactive prompts in non-interactive environments
& Poor handling of absent data and capability limits/out-of-scope cases: lacks fallbacks, conditional logic, or alternative suggestions \\
\addlinespace
Intent and channel selection errors: misinterprets the request, uses the wrong app/medium, or ignores medium-specific limits
& Missing required parameters or constraints (time, recipients, recurrence, labels, totals, etc.) \\
\addlinespace
Shopping cart and selection integrity issues: proceeds with contaminated carts, mishandles cart restoration or item-to-product mapping, and fails to validate product variants
& Insufficient handling of edge cases and input/format variations (e.g., boundary dates, file formats, catalog matching) \\
\addlinespace
Edge-case robustness deficiencies: mishandles time zones, date boundaries/rollovers, unit normalization, and variant-level inventory
& Missing finalization: fails to perform the final action (e.g., send/forward, purchase) or to return/deliver the final answer to the user \\
\addlinespace
Data integrity in inputs: uses hard-coded or unverified identifiers/values instead of extracting them from retrieved data
& Adds unsupported assumptions or extra details not provided by the user \\
\bottomrule
\end{tabular}
\caption{Top 10 system-level and node-level issues generated by Agentic CLEAR for the CUGA agent on AppWorld (GPT-4o backbone, GPT-5 judge), sorted by frequency. System-level analysis captures system-wide failure modes, while node-level analysis pinpoints component-specific root causes within the TaskDecompositionAgent.}
\label{tab:issues_system_node}
\end{table*}

%% file: latex/tables/trail_map_gpt5.tex
\begin{table*}[ht]
\centering
\footnotesize
\renewcommand{\arraystretch}{1.25}
\setlength{\tabcolsep}{4pt}
\begin{tabular}{@{}c p{0.43\textwidth} @{\hspace{8pt}} p{0.20\textwidth} @{\hspace{8pt}} p{0.24\textwidth}@{}}
\toprule
\textbf{\#} & \textbf{Issue} & \textbf{Full Match} & \textbf{Partial Match} \\
\midrule
1 & Did not leverage available tools and site-specific features; relied on memory or generic methods instead of invoking search/visit/inspect &
\emph{Tool Selection Errors} &
\emph{Hallucinations -- Lang.} \newline \emph{Poor Info.\ Retrieval} \\[4pt]

2 & Repeated tool misuse without correction (e.g., wrong arguments to functions, failure to use helper functions or archives) &
\emph{Tool Selection Errors} &
\emph{Hallucinations -- Tool} \newline \emph{Context Handling Failures} \newline \emph{Resource Abuse} \\[4pt]

3 & Made claims without case-specific substantiation: guessed/fabricated data, provided generic explanations, or asserted verification without reproducible evidence (quotes, permalinks, screenshots, tool outputs, URLs/parameters) &
\emph{Hallucinations -- Lang.} &
\emph{Hallucinations -- Tool} \newline \emph{Poor Info.\ Retrieval} \\[4pt]

4 & Failed to follow the outlined plan; skipped core retrieval and verification steps &
\emph{Task Orchestration} \newline \emph{Goal Deviation} &
\emph{Instruction Non-compl.} \\[4pt]

5 & Incomplete execution and coverage of the task: did not enumerate all candidates, apply filters, check all occurrences, or perform required computations/counts/filtering &
\emph{Task Orchestration} \newline \emph{Goal Deviation} &
\emph{Instruction Non-compl.} \\[4pt]

6 & Poor error recovery; repeated failing steps instead of pivoting to alternative strategies &
\emph{Task Orchestration} &
\emph{Resource Abuse} \newline \emph{Context Handling Failures} \\[4pt]

7 & Inefficient and redundant actions; excessive planning without tangible progress &
\emph{Resource Abuse} &
\emph{Task Orchestration} \\[4pt]

8 & Faulty or superficial parsing/extraction of source content (PDF/HTML or narratives), leading to incorrect values or miscounts &
\emph{Tool Output Misinterp.} &
\emph{Poor Info.\ Retrieval} \newline \emph{Formatting Errors} \\[4pt]

9 & Internal inconsistencies or logic errors within the analysis (e.g., off-by-one indices, contradictory counts, changing the required metric mid-analysis) &
NA &
\emph{Goal Deviation} \newline \emph{Incorrect Problem Id.} \newline \emph{Context Handling Failures} \\[4pt]

10 & Did not ensure requirements and scope clarity: failed to state/apply definitions, timeframe/version constraints, or to ask for clarification/pause when inputs were unavailable &
\emph{Incorrect Problem Id.} &
\emph{Instruction Non-compl.} \\[4pt]

11 & Output formatting and numerical precision requirements not followed (units, rounding rules, extra words, template violations) &
\emph{Formatting Errors} &
\emph{Instruction Non-compl.} \\[4pt]

12 & Minor factual inaccuracies and inconsistencies (e.g., misspelled names, wrong coordinates, misread code/grammar) &
\emph{Hallucinations -- Lang.} &
\emph{Tool Output Misinterp.} \\[4pt]

13 & Ignored explicit source requirements; did not consult the specified source &
\emph{Instruction Non-compl.} &
\emph{Poor Info.\ Retrieval} \\[4pt]

14 & Failed to verify data adjustment/measurement methodology (e.g., adjusted vs.\ unadjusted, intraday vs.\ close) &
\emph{Tool Output Misinterp.} &
\emph{Incorrect Problem Id.} \newline \emph{Poor Info.\ Retrieval} \\[4pt]

15 & Lack of cross-verification across independent sources &
\emph{Poor Info.\ Retrieval} &
\emph{Hallucinations -- Lang.} \\
\bottomrule
\end{tabular}
\caption{Mapping of GPT-5 system-level issues (GAIA) to TRAIL error categories. \emph{Lang.}\ = Language-only; \emph{Tool} = Tool-related; \emph{Misinterp.}\ = Misinterpretation; \emph{Id.}\ = Identification; \emph{Non-compl.}\ = Non-compliance; \emph{Info.}\ = Information. ``---'' indicates no full match.}
\label{tab:gpt5-trail-mapping}
\end{table*}

%% file: latex/tables/trail_map_120b.tex
\begin{table*}[ht]
\centering
\footnotesize
\renewcommand{\arraystretch}{1.25}
\setlength{\tabcolsep}{4pt}
\begin{tabular}{@{}c p{0.43\textwidth} @{\hspace{8pt}} p{0.20\textwidth} @{\hspace{8pt}} p{0.24\textwidth}@{}}
\toprule
\textbf{\#} & \textbf{Issue} & \textbf{Full Match} & \textbf{Partial Match} \\
\midrule
1 & Answers generated without retrieving or verifying data using tools, leading to unverified or fabricated information &
\emph{Hallucinations -- Lang.} &
\emph{Poor Info.\ Retrieval} \newline \emph{Tool Selection Errors} \\[4pt]

2 & Repeated identical or unnecessary tool calls &
\emph{Resource Abuse} &
\emph{Task Orchestration} \\[4pt]

3 & Poor error handling and lack of adaptation to failures &
\emph{Task Orchestration} &
\emph{Context Handling Failures} \newline \emph{Resource Abuse} \\[4pt]

4 & No source citations or evidence provided for answers &
\emph{Instruction Non-compl.} &
\emph{Hallucinations -- Lang.} \\[4pt]

5 & Misused tool arguments or invoked the wrong tool &
\emph{Tool Selection Errors} &
\emph{Hallucinations -- Tool} \\[4pt]

6 & Excessive planning steps that cause inefficiency &
\emph{Resource Abuse} &
\emph{Task Orchestration} \\[4pt]

7 & Missing required formatting tags or improper final answer handling &
\emph{Formatting Errors} &
\emph{Instruction Non-compl.} \\[4pt]

8 & Factual inaccuracies caused by unsupported assumptions &
\emph{Hallucinations -- Lang.} &
\emph{Poor Info.\ Retrieval} \\[4pt]

9 & Failed to handle missing or unsupported files gracefully &
\emph{Task Orchestration} &
\emph{Tool Output Misinterp.} \newline \emph{Context Handling Failures} \\[4pt]

10 & Did not use the most appropriate specialized API (e.g., Wikipedia API) for precise data extraction &
\emph{Tool Selection Errors} &
\emph{Poor Info.\ Retrieval} \\[4pt]

11 & Inefficient or incorrect handling of pagination and multi-page navigation &
\emph{Poor Info.\ Retrieval} &
\emph{Context Handling Failures} \newline \emph{Tool Output Misinterp.} \newline \emph{Resource Abuse} \\[4pt]

12 & Inappropriate tool selection for the task (e.g., omitting a required web\_search) &
\emph{Tool Selection Errors} &
\emph{Poor Info.\ Retrieval} \\
\bottomrule
\end{tabular}
\caption{Mapping of OSS-120B system-level issues (GAIA) to TRAIL error categories. Abbreviations as in Table~\ref{tab:gpt5-trail-mapping}.}
\label{tab:oss-trail-mapping}
\end{table*}